\documentclass{article}

\PassOptionsToPackage{numbers,sort&compress}{natbib}

\usepackage[preprint]{neurips_2026}
\workshoptitle{PTA: From Pretrained Representations to Acting Agents}

\usepackage[utf8]{inputenc}
\usepackage[T1]{fontenc}

\usepackage{microtype}
\usepackage{amsmath,amssymb,amsthm}
\usepackage{booktabs}
\usepackage{array}
\usepackage{tabularx}
\usepackage{multirow}
\usepackage{graphicx}
\usepackage{placeins}
\usepackage{float}
\usepackage{xcolor}
\usepackage{tikz}
\usetikzlibrary{arrows.meta,positioning,fit}
\usepackage{url}
\usepackage{hyperref}
\usepackage[nameinlink,capitalise,noabbrev]{cleveref}

\definecolor{frontierblue}{RGB}{39,88,132}
\definecolor{frontierlight}{RGB}{239,244,248}
\definecolor{frontiergreen}{RGB}{224,241,231}
\definecolor{frontierorange}{RGB}{250,235,215}
\definecolor{frontiergray}{RGB}{244,245,247}

\hypersetup{
  colorlinks=true,
  linkcolor=frontierblue,
  citecolor=black,
  urlcolor=frontierblue
}

\newtheorem{proposition}{Proposition}
\newtheorem{definition}{Definition}

\newcommand{\method}{\textsc{SayWhen}}
\newcommand{\acts}{\mathcal{A}}
\newcommand{\safeacts}{\mathcal{A}_{\mathrm{safe}}}
\newcommand{\particles}{\mathcal{Z}_t}
\newcommand{\massset}{\mathcal{I}_{\rho,t}}
\newcommand{\frontier}{\mathcal{F}_{\rho,t}}
\newcommand{\regret}{\mathcal{R}}
\newcommand{\indicator}{\mathbf{1}}
\newcommand{\release}{t_{\mathrm{rel}}}
\newcommand{\compat}{\preceq}
\newcommand{\valuecompat}{\preceq_{\mathrm{val}}}

\title{SayWhen: Set-Valued Semantic Commitment for Low-Regret Early Speech}

\author{%
  Tianyi Huang \\
  Ryquo
  \And
  Lucas Su \\
  App-In Club
  \And
  Fred Zhang \\
  App-In Club
  \And
  Andy Liang \\
  App-In Club
  \And
  Gordon Li \\
  App-In Club
  \AND
  \normalfont
  \texttt{tianyi@ryquo.com}\quad
  \texttt{\{luc, freddy, andy.liang, gordon\}@appinclub.org}
}

\begin{document}

\maketitle

\begin{abstract}
Speaking early turns a prediction into an irreversible public commitment. We introduce \method{}, a test-time controller that selects a task-directed act shared across typed continuation plans and controls its release as speech. A persistent veto blocks playback when intervening transcripts invalidate the candidate. We define regret-free early speech (RFES) to measure supported pre-endpoint releases without scored incompatibility. On a 72-prefix branching diagnostic, the generated-plan frontier attains 6.0\% RFES with zero observed mean regret, versus 8.2\% RFES and 0.068 regret for Matched Top-One. Measured planner latency constrains this path; a deterministic retail schema attains 31.7\% RFES, compared with 32.0\% for the complete policy. In a separate 30-task $\tau$-Voice simulation, schema-derived candidates with cached text-to-speech (TTS) audio attain 21/30 task-level RFES. Eligible audio begins 1.51 seconds before the user-audio endpoint among the 22 tasks with early speech, or 1.11 seconds over all 30 tasks with zero for tasks without eligible early speech. Semantic veto and endpoint waiting each pass 16/30 tasks. Together, the studies demonstrate a controlled coverage--compatibility trade-off and schema-based pre-endpoint operation, making the supported public commitment an explicit unit of voice-agent control.
\end{abstract}

\section{Introduction}

Natural conversation often leaves little silence between turns, in part because speakers begin planning before their interlocutor finishes \citep{stivers2009universals,levinson2015timing}. Conventional endpoint-triggered voice agents cannot exploit this overlap: they wait for endpoint detection and a final transcript before reasoning, after which speech synthesis adds further delay \citep{udupa2026endpoint}. Incremental dialogue systems instead interpret partial input and revise their state, while turn-projection and full-duplex models anticipate future speech activity or listen and speak concurrently \citep{schlangen2011incremental,baumann2011evaluation,ekstedt2022vap,defossez2024moshi}. Together, these capabilities raise a semantic control question absent from endpoint-triggered generation: \emph{what can a system safely make public while the user's turn remains unresolved?}\footnote{Companion demo: \url{https://github.com/Ryquo-Labs/SayWhen}.}

Consider the partial request ``Could you check order W1234567 and tell me whether~\ldots'' The user may continue with a question about tracking, return eligibility, exchange, or cancellation. Committing to one predicted suffix may force a repair; waiting forfeits the opportunity to reduce latency; a generic filler may add audio without identifying the task. Yet the branches share a useful act: retrieve the same order. The agent can begin that read-only work and say ``I'll pull up the order'' without resolving the downstream intent.

\begin{figure}[t]
  \centering
  \includegraphics[width=\linewidth]{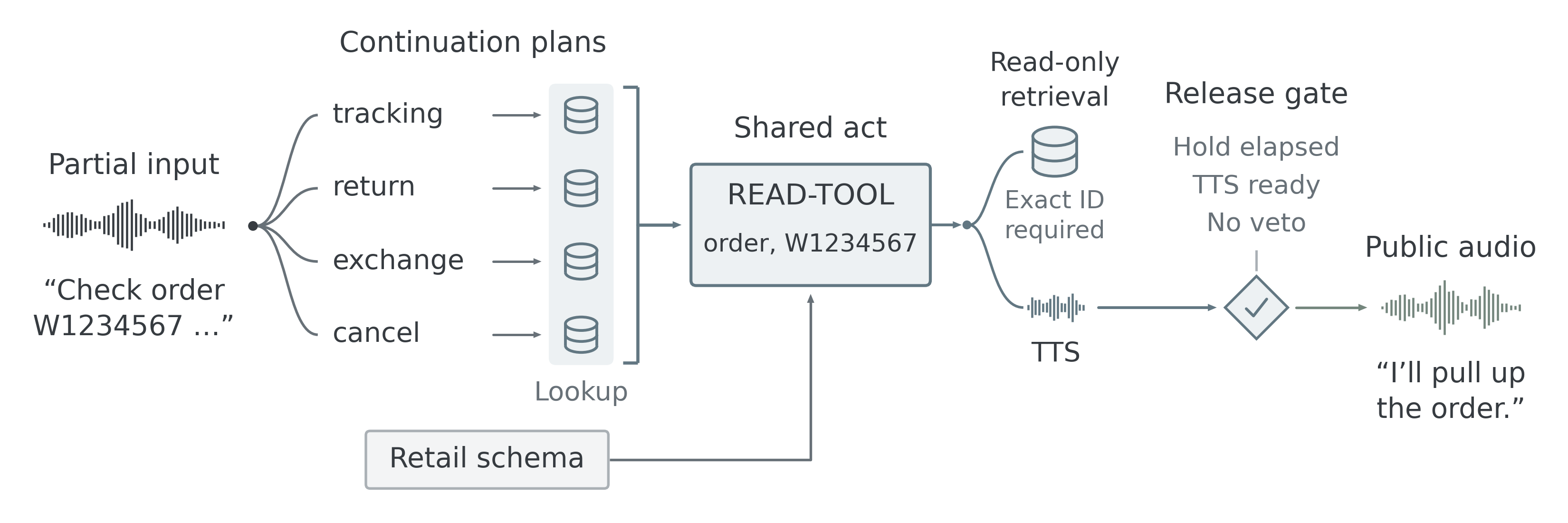}
  \caption{\textbf{SayWhen overview.}
    Four illustrative plans share a grounded order lookup despite different
    downstream intents. A retail schema provides an alternative candidate
    source and supplies all candidates in the online simulation. Retrieval
    and TTS proceed privately; the release gate controls public playback.}
  \label{fig:overview}
\end{figure}

\method{} implements this intersection of plans as a test-time interface between generated plans and public action (\cref{fig:overview}). We call its selector the \emph{Semantic Commitment Frontier}; it extracts a task-directed act shared across selected plans. A deterministic retail schema handles covered order requests without generative planning. Our \emph{Semantic Release Gate} controls public playback: retrieval and TTS may run privately, while speech requires the hold to have elapsed, TTS to be ready, and no intervening automatic speech recognition (ASR) update to have vetoed the candidate.

\paragraph{Contributions.}
(1) We formulate pre-endpoint speech as set-valued semantic commitment and define regret-free early speech (RFES), which requires at least one supported task-directed release before the endpoint and no scored pre-endpoint release incompatible with the realized interaction.
(2) We present a deterministic test-time frontier construction over generated typed plans, a targeted schema for predefined retail workflows, and a release gate with a persistent veto, together with explicit assumptions and abstention rules.
(3) We evaluate generated-plan selection in a branching diagnostic and schema-based release in a full-duplex retail simulation, separating semantic-selection effects from online task outcomes.

\section{Related work}

\paragraph{Incremental dialogue and early output.}
Incremental systems expose partial interpretations and revisions \citep{schlangen2011incremental}, learn whether to answer or wait from partial input \citep{devault2009canifinish,lu2011response}, and balance output quality against delay \citep{baumann2011evaluation}. Task-oriented systems have acted once an incremental semantic parse is stable enough \citep{ghigi2014incremental}; incremental interaction managers have also combined partial ASR with partially observable Markov decision process (POMDP) belief tracking, starting task-advancing actions and replacing them after later revisions \citep{selfridge2012integrating}. Incremental speech generation supports revisable plans, noncommitting segments, and output self-monitoring \citep{skantze2010generation}. Voice Activity Projection models predict turn shifts and backchannels \citep{ekstedt2022vap}, while full-duplex systems such as Moshi jointly represent simultaneous user and system streams \citep{defossez2024moshi}. Our setting imposes a distinct constraint: already-audible content cannot be revoked, so the model path selects a typed act supported across plausible continuations and controls its public release separately.

\paragraph{Plan-based dialogue and planning under uncertainty.}
Speech acts have long been modeled as planning operators \citep{cohen1979speechacts}, while generalized plan recognition maintains multiple possible plans rather than committing prematurely to one interpretation \citep{kautz1986planrecognition}. Other planning traditions address uncertainty through belief-state dialogue policies \citep{williams2007pomdp} or action sequences that remain valid across possible worlds \citep{smith1998conformant}. Planning landmarks are facts that must hold at some point in every solution plan \citep{hoffmann2004landmarks}. These traditions motivate conservative test-time action selection under uncertainty. Our object differs from a conformant plan or fact landmark: it is a typed public act compared across continuation-conditioned plans, with later ASR evidence able to cancel the candidate before audio release.

\paragraph{Overlapping computation.}
Online semantic parsing can execute programs predicted from prefixes \citep{zhou2022online}; response prefetching prepares answers for predicted utterances \citep{mori2025prefetching}; and Endpoint Anticipation forecasts turn completion so downstream work can begin early \citep{udupa2026endpoint}. SHANKS generates unspoken reasoning while listening and can make tool calls or interruption decisions before the user finishes \citep{chiang2026shanks}. These approaches move interpretation, reasoning, or action ahead of the endpoint. \method{} keeps such preparation private and asks when supported task-directed speech can be released.

\paragraph{Early public audio.}
Context-aware prefaces bridge response delay \citep{okafuji2026preface}; DDTSR produces minimal-commitment connectives while a larger model reasons \citep{liu2026ddtsr}; and RelayS2S uses a lightweight verifier to gate a speculative fast-path prefix before a slower continuation \citep{mai2026relays2s}. In contrast, \method{}'s model frontier searches for a typed act shared by every member of the selected continuation set, and the complete system treats candidate formation and public release as separate decisions. Each pre-endpoint release is evaluated against the realized interaction, while generic fillers do not qualify as task-directed early speech under RFES.

\paragraph{Streaming voice-agent evaluation.}
$\tau$-bench evaluates policy-constrained tool agents \citep{yao2025taubench}, $\tau^2$-Bench adds dual-control tasks in which both the user and agent take actions \citep{barres2026tau2}, and $\tau$-Voice combines grounded task completion with full-duplex audio \citep{ray2026tauvoice}. Full-Duplex-Bench-v2 evaluates multi-turn correction, entity tracking, and safety \citep{lin2026fdbv2}; the work-in-progress v3 adds tool use under disfluency and self-correction \citep{lin2026fdbv3}. Because a realized dialogue exposes only one continuation, we pair online simulation with a controlled suite that exposes sibling continuations of the same prefix.

\section{Early speech as semantic commitment}

At time $t$, the agent observes dialogue history $h_t$ and a partial transcript $x_t$; the remainder of the user turn is unknown. We represent each public commitment as a typed dialogue act $a\in\acts$. It records an act kind, a target, canonical keyed arguments such as identifiers or dates, and other canonical semantic arguments such as a requested detail or action scope. Canonicalization maps aliases and formatting variants to a shared value. An identifier, such as an order ID, determines which entity the act targets. Our act vocabulary includes \textsc{Frame-Task}, \textsc{Frame-Lookup}, \textsc{Read-Tool}, \textsc{Report-Fact}, \textsc{Ask-Clarification}, \textsc{Confirm-Action}, and \textsc{Write-Tool}. \textsc{Frame-Lookup} commits to a target-specific lookup workflow before an identifier is available; \textsc{Read-Tool} adds the grounded identifier and can license private retrieval. The implementation stores both under a common lookup-family label, which we report as \textsc{Frame-Lookup} when identifier-free and \textsc{Read-Tool} when grounded. Each releasable act maps to a prespecified utterance template whose wording must not add a new claim; any tool invocation is a separate private event.

Let $I(a)$ map identifier and date arguments to their canonical values, and let $\Phi(a)$ collect the remaining public propositions. Write $v\valuecompat w$ when $v=w$, or when $v$ is a year-omitted date and $w$ has the same month and day with a stated year. We write $a\compat b$ when their canonical kinds and targets agree, $\Phi(b)$ entails $\Phi(a)$, $\operatorname{dom}I(a)\subseteq\operatorname{dom}I(b)$, and $I(a)(q)\valuecompat I(b)(q)$ on every shared key $q$. This is a semantic preorder, not an ordering of plan prerequisites. The allowed act types $\safeacts\subset\acts$ exclude clarification, confirmation, and write acts. Let $E(P)$ be the order-preserving subsequence of plan acts retained for frontier construction. Candidate-level eligibility is time-dependent and enters separately through $\mathrm{Elig}_t$ in \cref{eq:common}. Grounding depends on the commitment: \textsc{Frame-Task} requires a recognized task; \textsc{Frame-Lookup} also requires a recognized target; \textsc{Read-Tool} additionally requires a canonical identifier; and \textsc{Report-Fact} requires the value to be available. Private retrieval always requires the identifier.

For user turn $u$, let $y_u$ denote its realized trace, including the completed continuation, dialogue context, endpoint $t_{\mathrm{eot},u}$, and the set of \emph{pre-endpoint} public releases
$S(y_u)=\{(a_j,t_j):t_j<t_{\mathrm{eot},u}\}$.
Here $t_j$ is the onset of public playback for act $a_j$. The two studies use different prespecified score ranges (Appendix~\ref{app:scoring}); we use $r(a_j;y_u)$ for the applicable deterministic incompatibility score. Public-commitment regret is
\begin{equation}
  \regret(y_u)=\max_{(a,t)\in S(y_u)} r(a;y_u),
  \qquad \max\varnothing=0.
  \label{eq:regret}
\end{equation}
Define the first RFES-eligible public onset by
\[
  t_m(y_u)=\inf\{t_j:(a_j,t_j)\in S(y_u),\ \mathrm{Elig}_{t_j}(a_j)=1\},
  \qquad \inf\varnothing=\infty.
\]
Then
\begin{equation}
  \begin{aligned}
  \mathrm{RFES}(y_u)
    &=\indicator[t_m(y_u)<t_{\mathrm{eot},u}]\,\indicator[\regret(y_u)=0],\\
  L(y_u)
    &=
  \begin{cases}
    t_{\mathrm{eot},u}-t_m(y_u), & t_m(y_u)<t_{\mathrm{eot},u},\\
    0, & \text{otherwise.}
  \end{cases}
  \end{aligned}
  \label{eq:rfes}
\end{equation}
The controlled diagnostic applies \cref{eq:rfes} to its single-turn branches. For online task $i$ with user turns $U_i$, let $y_{iu}$ be the realized trace of turn $u$ and define
\[
  M_i=\indicator[\exists u\in U_i:t_m(y_{iu})<t_{\mathrm{eot},iu}],
  \quad
  \regret_i=\max_{u\in U_i}\regret(y_{iu}),
  \quad
  \mathrm{TaskRFES}_i=M_i\indicator[\regret_i=0].
\]
Each evaluated task contains at least one user turn.
At either aggregation level, RFES requires at least one supported task-directed pre-endpoint act and no scored incompatibility across the relevant releases. It is a semantic-release measure: an eligible \textsc{Frame-Lookup} denotes a target-specific workflow commitment, not a completed retrieval; end-to-end completion is reported separately. Lead measures pre-endpoint overlap, not whether speaking at that instant is interactionally appropriate.

\section{\method{}}

\begin{figure}[ht]
  \centering
  \includegraphics[width=\linewidth]{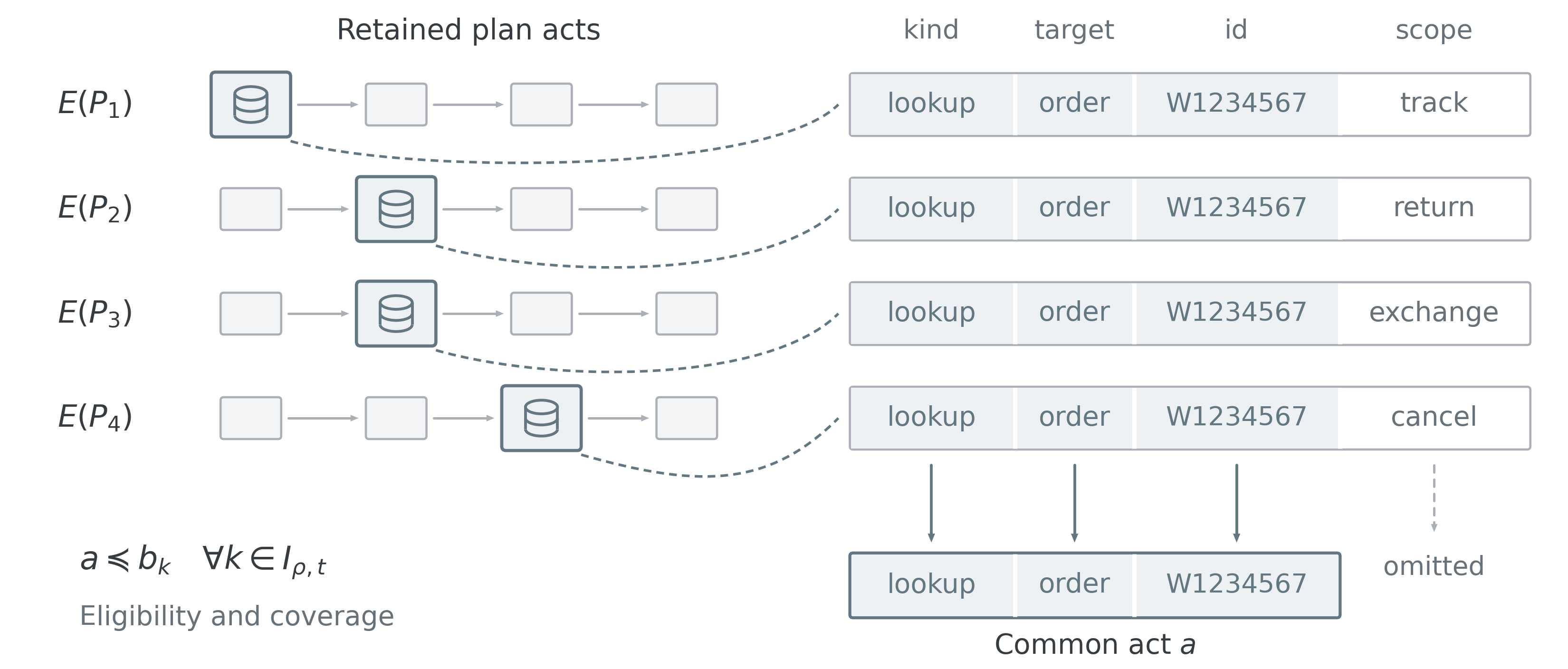}
  \caption{\textbf{Typed alignment and generalization.}
    Aligning the illustrated lookups $b_k$ and omitting their differing
    scopes yields a shared act $a \preceq b_k$ for every selected plan,
    preserving kind, target, and identifier. Pale cells denote other
    branch-specific acts.}
  \label{fig:saywhen-frontier}
\end{figure}

\subsection{Semantic Commitment Frontier}

The planner represents uncertainty using an ordered family of weighted continuation hypotheses
\begin{equation}
  \particles=(z_k)_{k=1}^{K},\qquad z_k=(c_k,p_k,P_k),
  \qquad p_k>0,\quad \sum_{k=1}^{K}p_k=1,
  \label{eq:particles}
\end{equation}
where $c_k$ is a plausible suffix and $P_k$ is its typed plan, compiled from a compact response frame. A single structured call returns all $K$ hypotheses and their positive self-reported weights. We normalize these weights within the batch; they are not calibrated continuation probabilities. For $0<\rho\leq 1$, let $\pi_t$ be a permutation of $\{1,\ldots,K\}$ that sorts hypotheses by decreasing weight using the fixed tie-breaks, and define
\begin{equation}
  m_{\rho,t}=\min\left\{m\in\{1,\ldots,K\}:\sum_{j=1}^{m}p_{\pi_t(j)}\ge\rho\right\},
  \qquad
  \massset=\{\pi_t(1),\ldots,\pi_t(m_{\rho,t})\}.
  \label{eq:mass-set}
\end{equation}
Equation~\eqref{eq:mass-set} defines the selected hypothesis set. The main setting uses $K=4$ and $\rho=1$, so every generated plan participates and the weights do not affect the main frontier. They still determine Matched Top-One and the lower-$\rho$ sweep.

Let $\mathrm{Elig}_t(a)\in\{0,1\}$ indicate whether an act remains supported, is task-directed, and satisfies its public semantic prerequisites at time $t$. In both studies, only \textsc{Frame-Task}, \textsc{Frame-Lookup}, \textsc{Read-Tool}, and \textsc{Report-Fact} can satisfy this predicate. Their prespecified utterances must convey the supported target or fact, either directly or through a logged deictic referent, and normalized fillers do not count (Appendix~\ref{app:scoring}). After normalizing equivalent surface forms and aligning acts by canonical kind, target, and grounded entity, the common commitment set is
\begin{equation}
  C_{\rho,t}(\particles)=
  \left\{
    a\in\safeacts:
    \mathrm{Elig}_t(a)=1\ \land\
    \forall k\in\massset\ \exists b\in E(P_k)\text{ with }a\compat b
  \right\}.
  \label{eq:common}
\end{equation}
The evaluated implementation deterministically returns an ordered frontier
\[
  \frontier(\particles)=(a_1,\ldots,a_m),
  \qquad \{a_i:1\leq i\leq m\}\subseteq C_{\rho,t}(\particles),
\]
whose order is supported by every selected plan. The constructor searches order-preserving alignments, prioritizing aligned length and then fixed specificity and serialization tie-breaks; consecutive duplicates are suppressed afterward. A final check rejects any sequence containing an ineligible act or one not covered by every selected plan. The policy uses the first frontier act as the turn's candidate, provided that no act has yet been released, and abstains when $m=0$.

For retail, a deterministic schema forms \textsc{Frame-Lookup}(order) when the partial transcript matches one of five predefined intents (track, return, exchange, cancel, or modify). If current-turn ASR yields an order identifier $d$, this candidate becomes \textsc{Read-Tool}(order, $d$). An exact identifier can also form the grounded candidate directly when the dialogue state already establishes the order-lookup task. Both paths remain subject to the checks below, and only the identifier-grounded form starts private retrieval. The complete controlled policy uses the schema for eligible retail cases and the model frontier otherwise; airline and telecom prefixes always use the model frontier. A \textsc{Report-Fact} is eligible only when its value is available; agreement on a field alone does not license its release.

\subsection{Semantic Release Gate}

\begin{figure}[t]
  \centering
  \includegraphics[width=\linewidth]{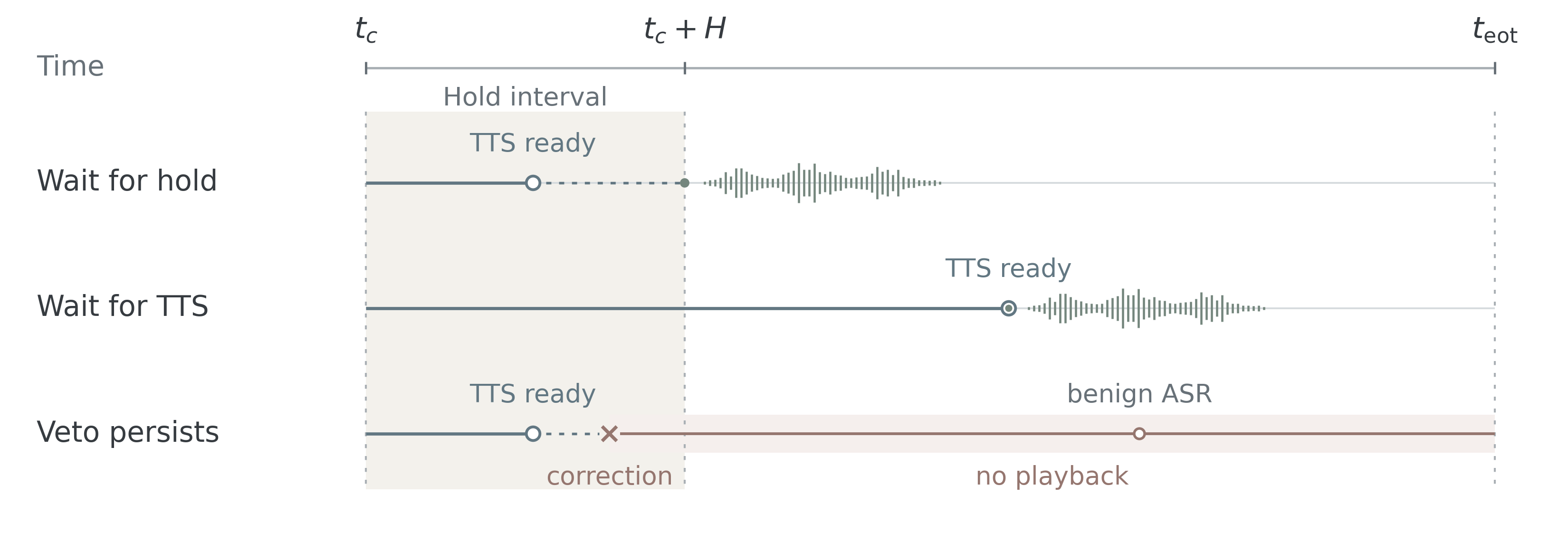}
  \caption{\textbf{Semantic Release Gate.}
    Shading marks the hold interval from candidate formation $t_c$.
    Solid slate segments show TTS preparation, open slate circles mark
    readiness, and dashes indicate private waiting. Green dots and waveforms
    mark public playback; concentric markers indicate simultaneous readiness
    and onset. A correction latches the veto despite later benign ASR
    updates. Timings are schematic.}
  \label{fig:saywhen-release-gate}
\end{figure}

An eligible act $a_c$ formed at time $t_c$ is a candidate, not immediate permission to speak. It is rejected if the current prefix already corrects or abandons the request, negates its reference, changes the entity, or contains a conflicting identifier. Otherwise, TTS and any licensed read-only retrieval may begin privately. The detector $g$ checks subsequent ASR evidence for these invalidations, including rolling revisions that remove the intent or replace its identifier. Let $j_c$ index candidate formation. With $e_c$ delimiting the formation evidence and $a_c^{(j)}$ allowing first-identifier enrichment, the veto evolves for subsequent ASR updates $j>j_c$ as
\begin{equation}
  v_j=v_{j-1}\lor g(a_c^{(j)},x_j;e_c),\qquad v_{j_c}=0.
  \label{eq:veto}
\end{equation}
A veto stays latched for the rest of the turn. ASR updates precede playback checks, including at a shared timestamp.

Let $q_{\mathrm{TTS}}(t)$ indicate that the audio is ready, and treat $q_{\mathrm{TTS}}$ and $v$ as right-continuous. Let $H\geq0$ be the minimum delay from candidate formation to public playback. Playback begins at
\begin{equation}
  \release=\inf\{t\in[t_c,t_{\mathrm{eot},u}):t-t_c\ge H,\ q_{\mathrm{TTS}}(t)=1,\ v(t)=0\}.
  \label{eq:release}
\end{equation}
We take $\inf\varnothing=\infty$; in that case, the candidate remains private. A matched blind hold uses the same candidate formation, $H$, and TTS timing but fixes $g\equiv0$ after formation. Immediate release sets $H=0$ and $g\equiv0$, opening playback as soon as the audio is ready. The policy releases at most one act per user turn, never revises released speech, and never executes or announces a state-changing operation pre-endpoint.

The first exact identifier can enrich an identifier-free candidate without restarting its hold or TTS. Online grounding scores use the final logged record, which may be enriched after playback, with a separate audible-content sensitivity analysis (Appendix~\ref{app:scoring}).

\section{Experimental design}

\begin{table}[ht]
  \caption{\textbf{Complementary evaluation settings.} The controlled diagnostic tests plan selection under shared timing; the online simulation tests schema-based release in full-duplex interaction.}
  \label{tab:evidence-map}
  \centering
  \small
  \setlength{\tabcolsep}{4pt}
  \begin{tabularx}{\linewidth}{@{}lXX@{}}
    \toprule
    & Controlled diagnostic & Online simulation \\
    \midrule
    Candidate source & Generated plans, retail schema, and combined policy & Retail schema only \\
    Model planning & One call per supplied prefix, returning $K=4$ plans & No frontier-planner calls \\
    Speech timing & Recorded planner and TTS intervals replayed & Fixed realization, cached TTS audio \\
    Evidence & Selection, coverage, and incompatibility & Playback timing and task outcomes \\
    \bottomrule
  \end{tabularx}
\end{table}

\subsection{Controlled branching diagnostic}

The controlled suite contains 144 programmatically constructed prefixes across airline, retail, and telecom domains. Each prefix has four author-specified continuations with fixed weights that sum to one, and each continuation is paired with a canonical gold plan. Six balanced strata cover explicit retrieval, implicit retrieval, no common eligible act, late correction, conflicting identifiers, and required clarification. We report the designated 72-prefix analysis partition (288 realized branches); the remaining 72 prefixes were reserved for method construction. This is a designed branching test, not a sample of naturally occurring conversations: branch weights specify the diagnostic mixture rather than estimated user frequencies.

The main configuration uses $K=4$, $\rho=1$, and $H=400$ ms. For each evaluated prefix, one structured \texttt{gpt-5.6-luna} call with reasoning disabled produces four hypotheses \citep{openai2026luna}; on the analysis partition, median planner latency is 1,599 ms (P90 1,966 ms). The supplied prefix is the planning trigger: the planner is invoked once, and its returned plans are held fixed during replay. Later ASR events update eligibility and the veto, without model replanning. Candidate formation, measured planner delay, the TTS startup interval, ASR event times, and the user-audio endpoint are replayed on a common event timeline. In the zero-planner-latency replay, planner completion, candidate formation, and the ensuing TTS request move earlier by the measured planner delay. The measured TTS startup interval is preserved, so TTS readiness advances by the same amount, while ASR-event and endpoint timestamps remain fixed. Public playback onset is determined by \cref{eq:release}; lead is the difference between the endpoint and that onset. Comparisons among top-one selection, the typed frontier, and first-act lexical agreement use exactly the same precomputed planner outputs. The complete policy also includes the retail schema, so its comparison with Matched Top-One is a system-level contrast rather than a one-factor ablation.

Endpoint waiting prohibits pre-endpoint speech. Matched Top-One selects the first task-directed act of an allowed type from the highest-weight hypothesis. First-act lexical agreement requires the fixed realizations of all selected plans' first task-directed acts of an allowed type to match after whitespace and case normalization. All generated-plan comparisons reuse the same hypotheses, release gate, planner delay, TTS timing, and event timeline.

The Gold-plan reference replaces generated hypotheses with all four author-specified continuations, their normalized branch weights, and canonical gold plans, then applies the same $\rho=1$ frontier. Its candidate is placed on the corresponding prefix's event timeline using the same recorded planner completion, TTS-ready event, ASR updates, and endpoint. The reference never conditions on the realized branch and retains measured planner timing, so it is a closed-world reference rather than a latency-free oracle. A deterministic typed matcher compares pre-endpoint releases with canonical gold plans; no human or model judge is used. Within each prefix, we use the fixed branch weights to average RFES, regret, and lead; we then average these prefix-level quantities equally. A branch without eligible pre-endpoint audio contributes zero lead.

\begin{table}[ht]
  \caption{\textbf{Controlled branching diagnostic.} Metrics are branch-weighted over four continuations per prefix and then averaged across the 72-prefix analysis partition. Matched Top-One and the $K=4$ typed frontier share generated plans and timing, isolating plan selection; the retail-schema and complete-policy rows show the coverage attribution. Unconditional lead is zero when no eligible speech begins before the endpoint.}
  \label{tab:controlled-main}
  \centering
  \small
  \begin{tabular}{lrrr}
    \toprule
    Method & RFES (\%) $\uparrow$ & Mean regret (0--3) $\downarrow$ & Uncond. lead (ms) \\
    \midrule
    Endpoint waiting & 0.0 & 0.000 & 0.0 \\
    Matched Top-One & 8.2 & 0.068 & 26.9 \\
    $K=4$ typed frontier & 6.0 & 0.000 & 16.5 \\
    \addlinespace
    Retail schema + gate & 31.7 & 0.000 & 301.2 \\
    Complete policy & 32.0 & 0.000 & 303.9 \\
    \addlinespace
    Gold-plan reference & 64.5 & 0.000 & 565.3 \\
    \bottomrule
  \end{tabular}
\end{table}

\subsection{Online full-duplex simulation}

We integrate the retail schema into the $\tau$-Voice tick loop \citep{ray2026tauvoice} and refer to its combination with the Semantic Release Gate as the \emph{semantic-veto} policy. Before evaluation, we used a fixed seed to draw a stratified set of 30 voice-enabled retail tasks (8 exchange, 8 return, 7 cancel, and 7 modify). We evaluate semantic veto, matched blind hold, immediate release, and endpoint waiting on every task, yielding 120 retained trajectories. The agent model is \texttt{gpt-realtime-2} \citep{openai2026realtime2}. All comparisons use the same simulator model and a fixed custom synthetic voice, enabling task-matched policy comparisons within this configuration.

For all three early policies, candidates come from the retail schema. The fixed acknowledgment is synthesized in advance, and its cached TTS audio is reused. Identifier-grounded candidates may also retrieve the order privately; identifier-free candidates do not. Semantic veto and matched blind hold use $H=400$ ms; the blind policy disables later vetoes, while immediate release uses $H=0$. Endpoint waiting opens no public playback before the user-audio endpoint.

We report Pass@1, the $\tau$-Voice binary completion score averaged over 30 tasks; task-level RFES; any-regret incidence $\indicator[\regret_i>0]$; and lead. Online RFES and regret use final-record grounding as a proxy for the release-time act; audible-only scoring is reported separately (Appendix~\ref{app:scoring}). Within each task, we first compute conditional lead by averaging $L(y_{iu})$ over user turns containing an eligible pre-endpoint release. The table then averages those task means over tasks with at least one such turn: 22 for semantic veto, 20 for matched blind hold, and 23 for immediate release. We also report an all-task lead: the same per-task conditional means, with zero for tasks without an eligible early release, averaged over all 30 tasks. This differs from a mean over all user turns. Paired percentile intervals resample task indices 10,000 times.

\section{Results}

With generated plans and timing held fixed, the $K=4$ typed frontier attains 6.0\% RFES with no observed public-commitment regret, compared with 8.2\% RFES and 0.068 mean regret for Matched Top-One (\cref{tab:controlled-main}). This matched comparison identifies an operating point with zero observed regret and a 2.2-percentage-point RFES trade-off. The corresponding 16.5 ms unconditional lead identifies planner speed as a practical bottleneck for generated-plan release.

The retail schema with the release gate attains 31.7\% RFES, nearly all of the complete policy's 32.0\%; the complete-policy comparison therefore reflects system-level coverage rather than the model frontier alone. The frontier's RFES rises from 6.0\% at measured latency to 22.2\% in the zero-planner-latency replay, indicating that planner latency constrains coverage in this setting. With measured timing matched, the 6.0\% generated frontier versus the 64.5\% Gold-plan reference diagnoses a separate gap in continuation or plan coverage and conservative constructor abstention.

\subsection{Plan selection and timing constraints}

First-act lexical agreement reaches 3.6\% RFES with 0.021 mean regret, versus 6.0\% RFES and 0.000 regret for the typed frontier (\cref{tab:structural-ablations,fig:controlled-overview}). This contrast combines semantic generalization with search beyond the first retained act; it does not isolate either change. The weight-threshold sweep retains the same zero-regret point estimate for the tested thresholds $\rho\in\{0.90,0.95,1.00\}$; lower thresholds expose a coverage--compatibility trade-off, modestly increasing RFES while introducing observed regret (\cref{tab:mass-sweep,fig:controlled-overview}).

\begin{table}[ht]
  \caption{\textbf{Online schema-based release.} One retained trajectory per task--policy pair on 30 fixed retail tasks. Outcome columns show counts out of 30; RFES and regret use final-record grounding. Lead is in milliseconds: conditional on tasks with eligible early speech, or averaged over all tasks with zero for no early speech. Neither lead measure is endpoint-to-answer latency.}
  \label{tab:online-main}
  \centering
  \small
  \setlength{\tabcolsep}{4pt}
  \begin{tabular}{@{}lrrrrr@{}}
    \toprule
    Policy & Pass@1 & Task RFES & Any regret & Cond. lead & All-task lead \\
    \midrule
    Semantic veto & 16/30 & 21/30 & 1/30 & 1507.6 & 1105.6 \\
    Matched blind hold & 14/30 & 19/30 & 1/30 & 1496.7 & 997.8 \\
    Immediate release & 18/30 & 23/30 & 0/30 & 1498.6 & 1148.9 \\
    Endpoint waiting & 16/30 & 0/30 & 0/30 & -- & 0.0 \\
    \bottomrule
  \end{tabular}
\end{table}

\subsection{Online simulation shows pre-endpoint operation}

Among the 22 semantic-veto tasks with at least one eligible pre-endpoint release, task-directed audio begins, on average, 1.51 seconds before the user-audio endpoint; assigning zero to the other eight tasks gives 1.11 seconds over all tasks (\cref{tab:online-main}). These lead summaries include eligible releases subsequently scored incompatible. The policy passes 16/30 tasks and meets primary task-level RFES on 21/30. The sole scored event was an ASR-induced final-record grounding mismatch; the identifier was unspoken, and audible-only scoring finds no incompatibility. Endpoint waiting also passes 16/30 tasks, yielding a paired Pass@1 difference of 0 percentage points with a 95\% interval of $-20$ to $20$ percentage points.

The semantic-veto policy's task-level RFES estimate is 6.7 percentage points above matched blind hold (95\% interval: 0--16.7 percentage points; \cref{fig:online-effects}); their conditional-lead estimates are similar at 1.51 and 1.50 seconds. Immediate release has the highest Pass@1 and task-level RFES point estimates and no observed regret in this run. Together, these runs demonstrate pre-endpoint schema operation; the paired intervals and single trajectory per task--policy pair leave the relative merits of the release policies unresolved.

\section{Discussion}

The results support treating commitment as a separate control decision from predicting a complete response. Generated plans expose acts shared across possible continuations, while a schema supplies candidates without a generative planner call for covered workflows. The measured and zero-planner-latency replays identify planner latency as a concrete target for improving the generated path. Keeping preparation private until release also provides a clear boundary between speculative work and the public commitments that RFES evaluates.

\paragraph{Interpreting responsiveness.}
RFES measures supported semantic release, whereas Pass@1 measures task completion. An identifier-free ``I'll pull up the order'' is a workflow commitment. Reported lead measures pre-endpoint audio overlap; endpoint-to-answer timing and human evaluation are needed to determine its effect on response speed and conversational usefulness.

\paragraph{Planning and speculative cost.}
The controlled study uses one structured call per prefix; the online schema reuses cached TTS audio without frontier calls. Evaluating repeated streaming replanning will require accounting for planning calls and tokens, private tool work, unused TTS, vetoed candidates, and endpoint-to-answer timing under the actual trigger policy.

\section{Scope and limitations}

The evaluation covers an authored three-domain diagnostic and a 30-task English retail simulation using models from one provider family. Broader conversational coverage, cross-family transfer, and perceived conversational quality remain open.

Conditional compatibility requires a selected plan that is sound for the realized turn, sound grounding, and a realization that adds no commitment beyond its typed act. Unseen continuations and missed corrections can still invalidate speech; zero observed regret is not a general guarantee. The release controller should be combined with turn-taking preferences and ordinary access controls. Speculative tools are read-only; state changes remain subject to resolved intent and required confirmation.

\section{Conclusion}

\method{} treats early speech as a public semantic commitment. Typed plan intersection yields a controlled coverage--compatibility trade-off, and a separate schema-based simulation demonstrates pre-endpoint release in full-duplex interaction. By separating candidate formation, private preparation, and public release, \method{} makes supported early speech a testable control objective while the user's request is still unfolding.

\begingroup
\small
\bibliographystyle{plainnat}
\bibliography{references}
\endgroup

\clearpage
\appendix
\raggedbottom

\section{\texorpdfstring{$\tau$}{tau}-Voice examples}
\label{app:logged-example}

\begin{table}[ht]
\caption{\textbf{Representative online decisions.} Retained $\tau$-Voice records illustrating an intent-grounded workflow commitment, identifier-grounded private retrieval on two separate user turns, and abstention. Negative times are relative to the logged user-audio endpoint; each tick is 200 ms. Routine dialogue and transport metadata are omitted.}
\label{tab:online-examples}
\centering
\small
\setlength{\tabcolsep}{3pt}
\renewcommand{\arraystretch}{1.15}
\begin{tabularx}{\linewidth}{@{}
  >{\raggedright\arraybackslash}p{19mm}
  >{\raggedright\arraybackslash}p{37mm}
  >{\raggedright\arraybackslash}p{40mm}
  >{\raggedright\arraybackslash}X@{}}
  \toprule
  Case & Logged evidence & Pre-endpoint behavior & Outcome \\
  \midrule

  \textbf{Intent-grounded commitment}\newline
  Task 52 &
  \textbf{Prefix} ($-2.0$ s, tick 35):
  ``Hi, I need to exchange\ldots''\newline
  \textbf{Complete}:
  ``Hi, I need to exchange a digital camera I bought.'' &
  The prefix licensed a \textsc{Frame-Lookup}(order)
  workflow commitment. No canonical identifier was available, so no
  private retrieval ran. After the 400 ms hold, the fixed lookup
  acknowledgment was released at $-1.6$ s. &
  \nolinkurl{exchange_delivered_order_items} remained gated by approval.
  The task passed, and the early release was regret-free. \\

  \addlinespace

  \textbf{Exact-ID private retrieval}\newline
  Task 113 &
  \textbf{User turn 6}: the rolling ASR hypothesis first contained the
  complete identifier \#W5056519 at $-2.2$ s; release at $-1.8$ s.\newline
  \textbf{User turn 8}: the rolling ASR hypothesis first contained the
  complete identifier \#W5995614 at $-1.8$ s; release at $-1.4$ s. &
  On these two separate user turns, each exact identifier licensed a successful private order retrieval
  before release of the fixed ``I'll pull up the order'' acknowledgment. &
  Both \nolinkurl{cancel_pending_order} calls remained gated by separate
  confirmations. The task passed; both releases were regret-free
  (mean lead: $1.6$ s). \\

  \addlinespace

  \textbf{Appropriate abstention}\newline
  Task 40 &
  One turn asks for a gift-card balance, the prior payment method,
  and a possible switch to Visa. &
  The schema formed no eligible pre-endpoint candidate. Account reads
  occurred later, and the payment change remained gated on confirmation. &
  After those reads and ``Yes, switch to Visa,''
  the \nolinkurl{modify_pending_order_payment} call executed. The task passed
  with no pre-endpoint release and therefore no scored regret. \\

  \bottomrule
\end{tabularx}
\end{table}

\FloatBarrier

\section{Additional controlled results}
\label{app:additional-controlled}

\begin{table}[ht]
  \caption{\textbf{Typed frontier versus first-act lexical agreement.} Both use the same generated hypotheses, release gate, and timelines. The frontier combines typed generalization with alignment over retained plan acts; first-act lexical agreement compares only each plan's first task-directed act of an allowed type. Unconditional lead is zero when no eligible speech begins before the endpoint.}
  \label{tab:structural-ablations}
  \centering
  \small
  \begin{tabular}{lrrr}
    \toprule
    Method & RFES (\%) & Mean regret (0--3) & Uncond. lead (ms) \\
    \midrule
    First-act lexical agreement & 3.6 & 0.021 & 14.5 \\
    Typed frontier & 6.0 & 0.000 & 16.5 \\
    \bottomrule
  \end{tabular}
\end{table}

\begin{figure}[ht]
  \centering
  \includegraphics[width=\linewidth]{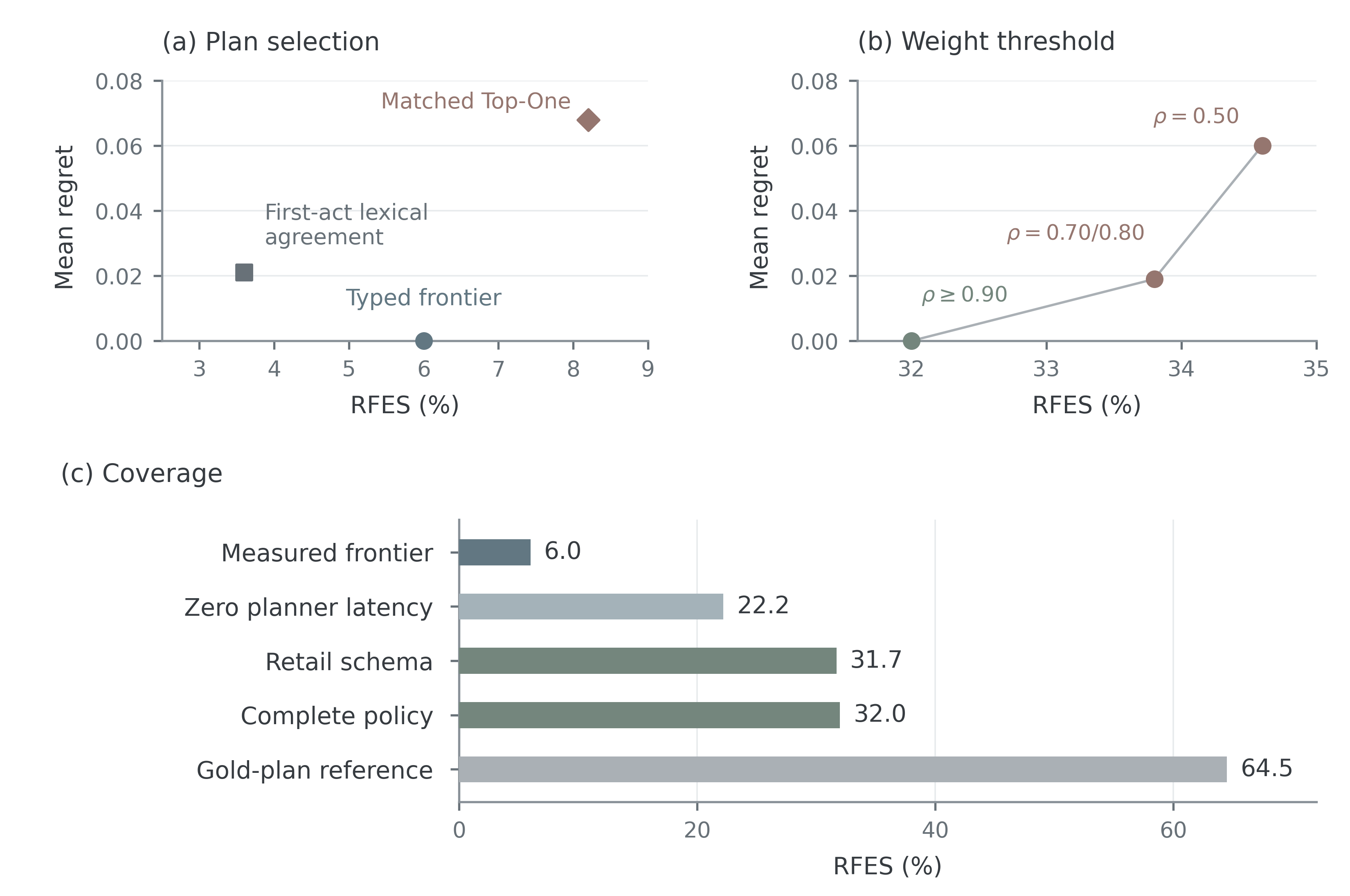}
  \caption{\textbf{Controlled operating points and coverage.}
    (a) Plan selection compares methods under shared generated plans and timing.
    (b) The threshold sweep varies $\rho$ within the complete policy; coincident
    settings are grouped. (c) Coverage bars compare distinct configurations, including
    a zero-planner-latency replay and a Gold-plan reference with measured latency.
    Values are branch-weighted estimates over the 72-prefix diagnostic.}
  \label{fig:controlled-overview}
\end{figure}

\begin{table}[ht]
  \caption{\textbf{Sensitivity to the cumulative-weight threshold $\rho$.} Only the model frontier's $\rho$ varies; generated hypotheses, the retail schema, release gate, and timing remain fixed. ``Incompatible branches'' denotes the branch-weighted percentage of realized continuations with an incompatible pre-endpoint release. Unconditional lead is zero when no eligible speech begins before the endpoint.}
  \label{tab:mass-sweep}
  \centering
  \small
  \begin{tabular}{lrrrrrr}
    \toprule
    $\rho$ & 0.50 & 0.70 & 0.80 & 0.90 & 0.95 & 1.00 \\
    \midrule
    RFES (\%) & 34.6 & 33.8 & 33.8 & 32.0 & 32.0 & 32.0 \\
    Mean regret (0--3) & 0.060 & 0.019 & 0.019 & 0.000 & 0.000 & 0.000 \\
    Incompatible branches (\%) & 3.0 & 1.0 & 1.0 & 0.0 & 0.0 & 0.0 \\
    Unconditional lead (ms) & 314.8 & 311.8 & 311.8 & 303.9 & 303.9 & 303.9 \\
    \bottomrule
  \end{tabular}
\end{table}

\FloatBarrier

\section{Online simulation details and scoring sensitivity}
\label{app:online-details}

The agent model is \texttt{gpt-realtime-2}; the user simulator is \texttt{gpt-4.1-2025-04-14}; rolling ASR uses \texttt{gpt-4o-mini-transcribe}. The semantic-veto policy permits one public act per user turn and uses a 400 ms hold, identifier-grounded private order retrieval, and cached TTS audio.

The final matrix contains 120 retained trajectories. Two attempts that failed because of infrastructure errors were rerun under an outcome-blind rule established before results were inspected.

The online loop has tick width $\Delta=200$ ms and rounds a duration $d$ upward to $\lceil d/\Delta\rceil$ ticks, so $H=400$ ms corresponds to two ticks. The simulator configuration contains a 400 ms hold and a 200 ms candidate-age check, both measured from candidate formation; their conjunction is exactly the 400 ms effective delay $H$ in \cref{eq:release}. At each tick, all available ASR events and veto updates precede the release check, and TTS becomes ready on the first subsequent tick after synthesis completes. Equation~\eqref{eq:release} states the continuous-time policy; the simulator applies this tick-quantized implementation.

The primary scorer treats a deictic expression and its grounded referent as one public semantic act. We additionally report an analysis that counts only incompatibilities audible in the waveform or stated in a post-release correction. Under this narrower rule, the semantic-veto policy obtains 73.3\% task-level RFES with 0\% audible incompatibility, whereas the primary analysis records a 3.3\% grounded-referent mismatch rate.

\begin{figure}[ht]
  \centering
  \includegraphics[width=\linewidth]{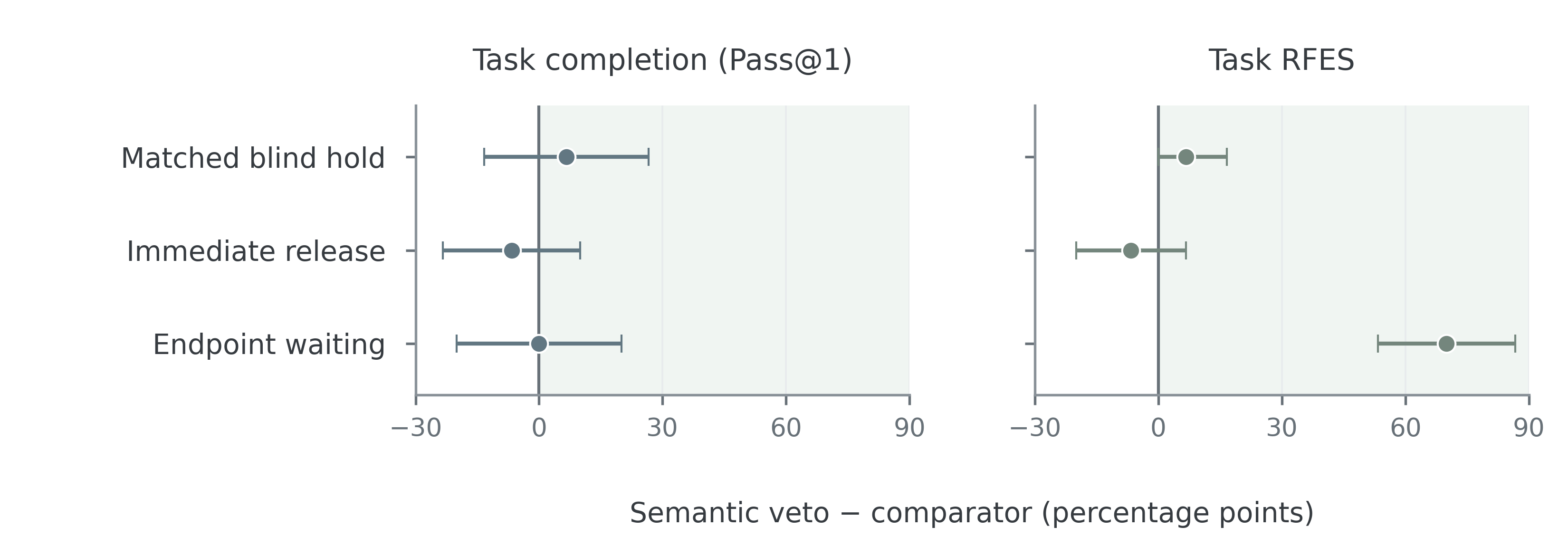}
  \caption{\textbf{Paired online-policy contrasts.}
    Differences are semantic veto minus comparator on the same 30 tasks.
    Points show Pass@1 (left) and task-level RFES (right); lines are 95\%
    percentile intervals from 10,000 paired bootstrap resamples over tasks.
    Endpoint waiting has zero RFES by construction, so its RFES contrast
    measures early-release coverage.}
  \label{fig:online-effects}
\end{figure}

\FloatBarrier

\begin{table}[ht]
  \caption{\textbf{Turn coverage and joint task outcomes.} Descriptive reanalysis of the same 120 retained trajectories. Early turns contain an eligible pre-endpoint release. Joint requires both task pass and primary final-record task-level RFES. Pooled lead weights every user turn equally, assigning zero without eligible early audio; it includes later-incompatible releases and is not endpoint-to-answer latency. Policies produce different trajectories and turn counts.}
  \label{tab:online-coverage}
  \centering
  \small
  \begin{tabular}{@{}lrrr@{}}
    \toprule
    Policy & Early / all user turns & Joint tasks & Pooled lead (ms) \\
    \midrule
    Semantic veto & 30/369 & 12/30 & 125.2 \\
    Matched blind hold & 33/363 & 8/30 & 138.3 \\
    Immediate release & 33/412 & 14/30 & 121.8 \\
    Endpoint waiting & 0/372 & 0/30 & 0.0 \\
    \bottomrule
  \end{tabular}
\end{table}

Table~\ref{tab:online-coverage} adds turn-level coverage and joint completion/RFES outcomes to the task-level results. The semantic-veto policy's 125.2 ms pooled lead is distinct from its 1105.6 ms all-task lead: the latter first averages only eligible-release turns within each task, then assigns zero to tasks without such speech. These summaries answer different weighting questions; neither estimates time saved.

\FloatBarrier

\section{Deterministic scoring rules}
\label{app:scoring}

\paragraph{RFES eligibility.}
A pre-endpoint release is eligible only if its logged prerequisites and grounding are satisfied. Its act must be \textsc{Frame-Task}, \textsc{Frame-Lookup}, \textsc{Read-Tool}, or \textsc{Report-Fact}, and its prespecified utterance must convey the supported target or fact directly or through its deictic referent. The online implementation approximates this grounding with the final candidate record, as detailed below. After lowercasing and removing terminal punctuation, the text must not match any of \{\texttt{got it}, \texttt{okay}, \texttt{ok}, \texttt{alright}, \texttt{sure}, \texttt{one moment}, \texttt{just a moment}\}. This defines $\mathrm{Elig}_t$; a log label alone is insufficient. An identifier-free logged \textsc{Read-Tool}(target) is reported as \textsc{Frame-Lookup}(target), which commits to the workflow but not an executed read. Private retrieval requires an exact canonical identifier.

\paragraph{Controlled severity.}
The controlled study sets $r(a;y)=r_{\mathrm{ctl}}(a;y)\in\{0,1,2,3\}$. Each release is greedily matched to the first remaining canonical gold-plan act $b$ with $a\compat b$; because policies release at most once per user turn, order is immaterial here. A matched release receives severity 0. Unmatched \textsc{Frame-Task}, \textsc{Frame-Lookup}, \textsc{Read-Tool}, and \textsc{Ask-Clarification} acts receive severity 2; unmatched \textsc{Report-Fact}, \textsc{Confirm-Action}, and \textsc{Write-Tool} acts receive 3; any other typed act receives 1. The total scale is defined over the full act vocabulary, although \method{} excludes several act types from early release. Trace regret is the maximum severity (\cref{eq:regret}); lead reports timing separately.

\paragraph{Online incidence.}
The online study approximates the release-time act using the final logged candidate and sets $r(a;y)=r_{\mathrm{on}}(a;y)\in\{0,1\}$ for this record-grounded proxy. The scorer compares the final logged candidate record with the completed transcript and recorded correction events. Identifier enrichment can update that record after playback, so this primary grounding score is not a release-time snapshot annotation; the audible-only sensitivity excludes unspoken identifiers. The scorer sets $r_{\mathrm{on}}(a;y)=1$ if the completed trace conflicts with the released act's grounded identifier, abandons or corrects the request, or changes the entity, and sets it to 0 otherwise. Detector events aid diagnosis but do not determine regret. Task aggregation uses the task-level definition immediately following \cref{eq:rfes}; ``Any regret'' in \cref{tab:online-main} is $\indicator[\regret_i>0]$, and the two studies' score ranges are never pooled.

\section{Diagnostic records and evaluated planning schedule}
\label{app:protocol}

\begin{table}[ht]
  \caption{\textbf{Controlled diagnostic specification.} The full suite has 144 prefixes; results use 72 designated analysis prefixes and their four branches. These fields describe the evaluated protocol, not a natural conversation distribution.}
  \centering
  \small
  \begin{tabularx}{\linewidth}{@{}lX@{}}
    \toprule
    Component & Specification \\
    \midrule
    Domains & Airline, retail, telecom \\
    Designed strata & Explicit retrieval; implicit retrieval; no common eligible act; late correction; conflicting identifiers; required clarification \\
    Prefix record & Partial transcript, four author-specified suffixes, fixed branch weights summing to one, canonical gold plan per branch \\
    Replay inputs & Recorded planner completion, TTS startup interval, ASR event times, and user-audio endpoint \\
    Aggregation & Weighted average across each prefix's branches, then equal average over analysis prefixes \\
    \bottomrule
  \end{tabularx}
\end{table}

\paragraph{Construction and partitioning.}
The original builder enumerates eight prefix variants for each of the three domains and six strata. Within each domain--stratum cell, variant indices 0--1 form calibration, 2--3 development, and 4--7 the analysis partition, giving 36, 36, and 72 prefixes respectively. Seed 20260808 shuffles record order without changing this assignment. The raw builder estimates continuation duration from word counts; the reported timed replay uses the recorded speech and event timelines described in Section~5.1.

The source supplement includes the 144 authored input records, canonical branch plans and weights, and the split manifest. Its raw duration fields are construction estimates, separate from the measured replay timelines. These inputs and the retained online-summary analysis support inspection of the diagnostic and reported aggregates; they do not include the complete audio, ASR, and generated-plan replay artifacts.

\paragraph{One batch per supplied prefix.}
Each controlled planning decision uses one structured call returning $K=4$ suffixes, weights, and response frames; typed plans are compiled from those frames. Later replay events can invalidate the candidate but do not request new plans. This protocol evaluates a decision at a fixed prefix, not a live trigger or repeated-replanning policy. Online candidates instead come from the retail schema as partial transcripts arrive. Before playback, the first exact identifier may enrich a candidate without restarting its hold; a grounded identifier is never replaced, and a vetoed candidate is not replaced within that turn. Available ASR and veto updates precede each playback check.

\paragraph{Evidence offsets and persistent state.}
The detector checks the transcript after stored offset $e_c$ for correction, abandonment, negation, or entity and identifier changes. Rolling revisions can also invalidate earlier intent or grounding. First-identifier enrichment preserves the target and does not trigger an identifier-change veto; a conflicting or replacement identifier does. ASR updates precede release checks; $v(t)$ and TTS readiness are treated as right-continuous.

In the online implementation, first-identifier enrichment updates the candidate in place, even after playback. It preserves $t_c$, $e_c$, the hold, and any latched veto; prepared audio is reused, and grounded private retrieval may begin. A vetoed candidate is not replaced. The fixed acknowledgment is unchanged; primary scoring reads the final record (Appendix~\ref{app:scoring}).

\paragraph{Weight validation and ordering.}
Every retained suffix must be nonblank and its raw self-reported weight $\tilde p_k$ finite and strictly positive. A missing, malformed, zero, negative, or nonfinite weight invalidates the planner attempt rather than being clipped or imputed. Normalization uses $p_k=\tilde p_k/\sum_j\tilde p_j$. Hypotheses are sorted by decreasing weight, then lexicographically decreasing case-folded suffix, then increasing original index. At $\rho=1$, all four plans participate regardless of their weights. Matched Top-One uses the same weight and suffix ordering, preserving original batch order on a full tie.

\paragraph{Baseline construction.}
Endpoint waiting prohibits pre-endpoint speech. Matched Top-One uses the hypothesis ordering above and takes the highest-weight plan's first task-directed act of an allowed type, abstaining if absent. First-act lexical agreement applies the fixed utterance template to each selected plan's first such act; it abstains if any act is missing or the whitespace-normalized, case-folded strings differ. Otherwise, it uses the act from the first selected hypothesis and common realization. Unlike the typed frontier, it does not align later plan acts. All generated-plan variants share the gate, measured planner latency, TTS timing, and event timeline.

\paragraph{All-task lead derivation.}
Let $\bar L_i$ be the mean eligible-release lead within task $i$, with $\bar L_i=0$ when task $i$ has no such release. Let $n_M=\sum_i M_i$. For $n_M>0$, the conditional and all-task summaries satisfy
\[
  \bar L_{\mathrm{cond}}=\frac{1}{n_M}\sum_{i:M_i=1}\bar L_i,
  \qquad
  \bar L_{\mathrm{all}}=\frac{1}{30}\sum_i\bar L_i
  =\frac{n_M}{30}\bar L_{\mathrm{cond}}.
\]
When $n_M=0$, conditional lead is undefined and all-task lead is zero. Table~\ref{tab:online-main} uses $n_M=22,20,23$ and reported conditional means, inheriting their rounding. Both summaries include later-incompatible releases. Endpoint-to-answer latency and duration-normalized lead cannot be inferred from these aggregates.

\section{Interpreting the online gate comparisons}
\label{app:gate-diagnosis}

The common task set matches task identities, not the ASR events or candidate opportunities encountered after different policies speak. The 21/30 versus 19/30 task-level RFES comparison is therefore a task-matched policy contrast, not a count of prevented errors. Immediate release's higher point estimates likewise do not establish a policy ranking.

As a timing consistency check on the saved semantic-veto summaries, candidate formation plus the recorded 400 ms effective hold reconstructs all 30 recorded pre-endpoint onsets on the 369 retained user turns, assuming the cached TTS audio is ready. The summaries lack the full ASR and invalidated-candidate history needed to estimate prevented errors. A candidate-matched gate evaluation would hold those events and actual TTS readiness fixed, then measure incompatible releases suppressed, supported opportunities lost, and lead.

\FloatBarrier
\clearpage

\section{Companion demonstration}
\label{app:demo}

The companion repository (\url{https://github.com/Ryquo-Labs/SayWhen}) provides a demonstration of private preparation and release checks. It intersects free-form answer openings and releases after the endpoint and final transcript; it does not implement the paper's typed pre-endpoint evaluations. Its scenarios and latency displays are illustrative, and the repository contains demonstration code rather than the full experimental pipeline.

\section{Formal definition and conditional compatibility}
\label{app:formal}

\begin{definition}[Semantic commitment preorder]
For canonical values, $v\valuecompat w$ iff $v=w$, or $v$ is a canonical date without a year and $w$ has the same month and day with a stated year. Map \textnormal{\textsc{Frame-Lookup}} and \textnormal{\textsc{Read-Tool}} to one canonical lookup kind. For typed acts $a,b\in\acts$, $a\compat b$ iff their canonical kinds and targets are identical, $\Phi(b)$ entails $\Phi(a)$, $\operatorname{dom}I(a)\subseteq\operatorname{dom}I(b)$, and $I(a)(q)\valuecompat I(b)(q)$ for every $q\in\operatorname{dom}I(a)$. These component relations are reflexive and transitive. Semantically equivalent acts can remain syntactically distinct, so $\compat$ is a preorder rather than a partial order. Time-dependent prerequisites and grounding are handled separately by $\mathrm{Elig}_t$.
\end{definition}

\paragraph{Frontier construction and date handling.}
Plan records store canonical keyed arguments in $J(a)$. Identifier and date entries form $I(a)$; other entries and the record's scope, field, slot, and content instantiate $\Phi(a)$. In plan order, $E(P)$ retains lookup-family, \textsc{Frame-Task}, and non-null \textsc{Report-Fact} acts from among the first seven plan acts; the cap is inactive because retained plans have at most five acts. Let $\ell(P)=|E(P)|$.

Tuple-wise generalization requires identical canonical kinds and targets. \textsc{Frame-Task} and lookup-family acts require equal $J$-domains, retaining only equal keyed values and scopes. For dates sharing a month and day, conflicting years are rejected and a year is retained only if every input states it. \textsc{Report-Fact} requires exact arguments, scope, field, and content; a tuple failing these conditions is undefined. Every output must lie below each input under $\compat$. The evaluated binary fold instead retains a uniquely stated year, but the final coverage check rejects that stronger output. No reported alignment mixes year-specific and year-omitted dates, so the results are unaffected.

Requiring equal $J$-domains makes the constructor a conservative search over $C_{\rho,t}$; the final check rejects any act stronger than an aligned input.

An act is serialized by kind, target, sorted $J(a)$, scope, field, slot, and content; its specificity is $2|J(a)|$ plus one per nonempty target, scope, field, slot, or content. The selected plans $\{P_k:k\in\massset\}$ are ordered by increasing $\ell(P)$ and total specificity, then lexicographically increasing serialization; the first is the anchor. From the anchor's longest subsequences downward, the constructor enumerates order-preserving alignments in every other plan and folds typed generalization over each aligned tuple. At the first feasible length, it chooses the greatest total specificity and then the lexicographically greatest serialized sequence. It suppresses consecutive duplicates and verifies coverage. Thus, for each $k\in\massset$, there are indices
\[
  1\leq q_{k,1}<\cdots<q_{k,m}\leq |E(P_k)|
  \quad\text{such that}\quad
  a_i\compat E(P_k)_{q_{k,i}}\quad\text{for all }i.
\]
The search maximizes aligned length before duplicate suppression; it does not claim to return a globally longest released frontier or an optimal dialogue plan.

\begin{proposition}[Conditional frontier compatibility]
Call $P^*$ sound for the realized trace $y$ if every act in $E(P^*)$ is semantically licensed by $y$. Suppose $P_k=P^*$ for some $k\in\massset$, $a\in C_{\rho,t}(\particles)$, compatibility and canonicalization are sound, semantic licensing is closed under weakening by $\compat$, and the utterance template adds no commitment beyond $a$. Then realizing $a$ adds no public proposition or keyed commitment beyond what $P^*$ licenses.
\end{proposition}

\begin{proof}
Because $P^*$ is selected, \cref{eq:common} gives $b^*\in E(P^*)$ with $a\compat b^*$. Soundness and closure under semantic weakening license $a$ without adding a proposition or keyed value beyond $b^*$. Membership in $C_{\rho,t}$ supplies eligibility and the act's time-dependent prerequisites; the template adds nothing further.
\end{proof}

\end{document}